\documentclass[letterpaper]{article} 
\usepackage{aaai25}  
\usepackage{times}  
\usepackage{helvet}  
\usepackage{courier}  
\usepackage[hyphens]{url}  
\usepackage{graphicx} 
\usepackage{natbib}  
\usepackage{caption} 
\usepackage{algorithm}
\usepackage{algorithmic}
\usepackage{amsmath}
\usepackage{array}
\usepackage{dsfont}
\usepackage{enumitem} 
\usepackage{multirow}
\usepackage{amssymb}
\usepackage{filecontents}
\usepackage{amsthm}
\usepackage{tikz}
\usepackage{tikz}
\usepackage{pgfplotstable}
\usetikzlibrary{positioning}
\usetikzlibrary{pgfplots.groupplots}
\usetikzlibrary{arrows}
\usepackage{pgfplots,mathcomSTEv4}
\usepackage{accents}

\usepackage{subfig}
\usepackage{amsfonts}
\usetikzlibrary{decorations.pathreplacing}

\usepackage{newfloat}
\usepackage{listings}
\DeclareCaptionStyle{ruled}{labelfont=normalfont,labelsep=colon,strut=off} 
\floatstyle{ruled}
\newfloat{listing}{tb}{lst}{}
\floatname{listing}{Listing}
\usepackage{color}

\newcommand{\topK}{ {\mathsf{top}}_{\kappa}}

\newtheorem*{assumption*}{Assumption}
\newtheorem{remark}{Remark}

\usepackage{tikz}
\usetikzlibrary{shapes, arrows.meta, positioning}
\usepgfplotslibrary{fillbetween}

\title{
Age-of-Gradient Updates for Federated Learning over \\ Random Access Channels
}

\author{
    Yu Heng Wu\textsuperscript{\rm 1}, 
    Houman Asgari\textsuperscript{\rm 2}, 
    Stefano Rini\textsuperscript{\rm 3}, 
    Andrea Munari\textsuperscript{\rm 4}
}
\affiliations {
    yuhengwu.ee13@nycu.edu.tw, houman.asgari@tum.de, stefano.rini@nycu.edu.tw, andrea.munari@dlr.de
}

\begin{document}

\maketitle

\begin{abstract}
This paper studies the problem of federated training of a deep neural network (DNN) over a random access channel (RACH), commonly used in wireless and sensor networks. 
More precisely, we consider the scenario in which a set of remote users participate in training a centralized DNN model using stochastic gradient descent (SGD) with the support of a parameter server (PS).
The local model updates are transmitted from the remote users to the PS over a RACH using a slotted ALOHA protocol, prone to packet losses due to interference. 
The PS collects the updates from the remote users that have been successfully received, accumulates them, and sends a model update to the users at regular time intervals. 
We refer to this setting as the RACH-FL setting. 

The RACH-FL setting crucially addresses the problem of jointly designing (i) a client selection and (ii) a gradient compression strategy which tackle the communication constraints between the remote users and the PS when transmission occurs over a RACH.
For the RACH-FL setting, we propose a policy, which we term ''age-of-gradient'' (AoG), in which (i) gradient sparsification is performed using $\topK$ sparsification, (ii) the slot transmission probability is obtained by comparing the current local memory magnitude and the global gradient updates, and (iii) the errors arising from both compression and transmissions are mitigated through error feedback.
Intuitively, the AoG policy measures the ''freshness'' of the local gradient and is used by the remote users to decide their transmission probability over the slotted ALOHA protocol. The solution we propose is somewhat inspired by  of the concept of age of information (AoI) in the context of communication theory, which measure how up-to-date information available at a terminal is. 
%
Numerical simulations show the superior performance of the AoG policy as compared to other RACH-FL policies. 
%
%
\end{abstract}

\section{Introduction}
\label{sec:Introduction}

Random Access Channels (RACHs) are a key component of various types of networks, including computer, wireless, and sensor networks, allowing terminals to transmit information over a shared medium without coordination, as epitomized by the slotted ALOHA protocol \cite{Abramson77_PacketBroadcasting}.
%
%
The approach is particularly well-suited for scenarios where a large number of users need to sporadically access a communication medium to interact with a central server,  without undergoing formal enrollment processes. 
It facilitates efficient data transmission and access in dynamic environments where users may join or leave the network frequently.
For this reason, it represents the standard approach in communication theory for applications requiring flexible and scalable solutions.

In this paper, we introduce a formal definition of a federated learning (FL) problem in which communication occurs over a RACH using slotted ALOHA -- which we term the RACH-FL setting. 
We discuss the fundamental trade-offs between training and communication  and argue that designing joint training/communication strategies for the RACH-FL setting provides a compelling framework with significant relevance for developing practical distributed learning protocols.
We then propose a practical strategy for the RACH-FL setup, named the "Age-of-Gradient" (AoG) policy, which is inspired by the Age of Information (AoI) concept from communication theory. This policy enables scalable decision-making for users on whether to transmit over the RACH based on the freshness of their local gradient information.

Our analysis demonstrates that this policy is not only theoretically robust but also immediately applicable. 
The paper concludes by highlighting that this model lays the groundwork for developing advanced protocols that are both practical and adaptable, addressing real-world challenges in distributed learning with improved performance and efficiency.

\medskip

Before delving into the main content, it is useful to preview some key trade-offs that this paper addresses.

Consider a system where communication between \( U \) remote users and the parameter server (PS) occurs over a RACH in which the successful transmission of a message allows delivery of \( \Rsf \) units of information (floats) per time unit (sec).
The PS updates the central model at regular intervals of \( T \) time units, as illustrated in Fig.~\ref{fig::FL Setting} and Fig.~\ref{fig:RACH setting}. The system designer has the flexibility to configure the slotted ALOHA protocol by choosing \( K \) communication slots between consecutive model updates.
For a given \( K \), each slot can accommodate \( \Rsf T/K \) units of information. Under the slotted ALOHA protocol, each remote user independently attempts transmission in any given slot with  probability $p$.
Leaning on the well-established collision channel model \cite{Abramson77_PacketBroadcasting}, if only one user transmits in a slot, the transmission is successful. However, if multiple users transmit in the same slot, a collision occurs, leading to communication failure.
Accordingly, a channel access probability $p=1/U$ maximizes the \emph{throughput} of the RACH \cite{Abramson77_PacketBroadcasting}, defined as the fraction of slots resulting in successful transmissions. 
%
%
Since the communication rate is constant, the remote users must compress their gradients to $\Rsf T/K$ units of information before transmission.
%
%
%
%
The parameters of the RACH channel, namely \(\Rsf\), \(T\), and \(K\), can be assumed to be fixed, given that the communication infrastructure must support numerous processes simultaneously.
%


Given the settings described above, we can adopt the perspective of the learning architect: our objective is to devise a scalable policy that regulates the number of users actively attempting transmission-- the  \emph{active users}-- to achieve efficient training while managing the inherent randomness associated with the channel transmission.
%

As we shall demonstrate, the results derived in this paper match the practitioner's wisdom -- see Fig.~\ref{fig: active users vs accuracy} --, in that (i) a higher number of active users is beneficial at the beginning of training, while (ii) a lower number of active users is preferable in the later stages. 
This can be interpreted as follows: (i) during the initial training phase, the model undergoes rapid changes, and having more active users helps average out gradient variability. 
Conversely, (ii) in the latter part of training, reducing the number of active users can enable
the PS to more reliably gather crucial gradient contributions from users with difficult-to-classify dataset entries.

%
%

Having established how the optimal number of active users varies throughout the training process, the next challenge is to develop a scalable policy that approximates this optimal choice. 
Specifically, we aim for this policy to rely solely on information available at the remote users, minimizing the need for coordination with the parameter server and other users. 
%
%
%
To address this, we propose the Age-of-Gradient (AoG) policy, which is inspired by the concept of ``freshness of information'' in communication theory \cite{yates2021age} and utilizes both global and local gradient information to implement a simple threshold-based decision rule for determining whether a user should be active \textit{(see Fig.~\ref{fig:threshold policy}, ~\ref{fig:threshold policy k=5})}.

\vspace{1cm}

\section{Related Work}
\label{sec:Related Work}

The role of communication constraints in FL has been extensively studied in the literature. Various gradient compression techniques \cite{dean2012large, mcmahan2017communication, wen2017terngrad, Shalev-Shwartz2010FL_CE, Wang2018Spars_FL, Alistarh2018Spars_FL, Bernstein2018signSGC, FL_DSGD_binomial, Li2019DP_CEFL} have been proposed to mitigate the computational burden of transmitting model updates from remote users to the PS. These approaches typically assume a noiseless, rate-limited channel between the users and the PS, focusing primarily on reducing the size of local updates. For example, \(\topK\) sparsification \cite{alistarh2017qsgd, wangni2018gradient} retains only the gradient coordinates with the largest magnitudes, while other methods use low-precision quantization, limiting the precision of gradient entries to as few as one \cite{Bernstein2018signSGC} or two bits \cite{QSGD} per entry. Additionally, error correction techniques, such as error feedback, have been shown to enhance performance \cite{seide2014onebitSGD, karimireddy2019error, stich2018sparsified}.

In addition to gradient compression, reducing the communication burden can be achieved by selectively choosing a subset of users for transmission. Early work explored random client selection to mitigate the impact of heterogeneous datasets \cite{li2020federated}. Later, methods such as client selection based on local loss \cite{goetz2019active} and the "power of choice" strategy \cite{cho2020client} aimed to select clients that provide the greatest improvement to model accuracy. More recent work \cite{ribero2020communication} proposed a client selection criterion based on the progression of clients’ weights.

While these approaches primarily assume a noiseless communication channel subject to rate constraints, they do not address the intrinsic contention for communication resources, where collisions can occur if multiple users attempt to transmit simultaneously. This aspect is crucial for scenarios where access to the communication medium is contention-based, necessitating careful design of the access policy to avoid collisions and ensure efficient learning.

For more challenging channel models, insights can be drawn from the wireless communications community. For instance, \cite{afsane2020} examines a slotted ALOHA medium access control protocol for uplink communication, proposing an iteration-termination criterion that balances training performance with computational, communication, and energy costs. Similarly, \cite{yang2019scheduling} addresses optimizing device participation in FL, providing tractable expressions for the convergence rate in a wireless setting, considering scheduling schemes and inter-cell interference. \cite{wang2024convergenc} introduces a Stackelberg game-based approach to jointly optimize global loss and latency, incorporating the concept of age-of-update (AoU) to prioritize devices that have not participated recently, thereby improving the freshness of information aggregated by the server.

%
Overall, while these studies provide valuable insights, they often involve complex channel models or performance optimization problems that may limit the applicability of their results across different scenarios. 
In particular, none of these models explicitly address the trade-off between the number of active users and learning performance, a trade-off that naturally arises in uncoordinated systems. This paper aims to address this gap by highlighting and analyzing this trade-off.

%

\smallskip
\noindent
{\bf Notation.}
Lowercase boldface letters (e.g., $\zv$) are used for tensors, 
whereas calligraphic uppercase for  sets (e.g. $\Acal$).
Given the set $\Acal$, $|\Acal|$ indicates its cardinality. 
We also adopt the short-hands  $[m:n] \triangleq \{m, \ldots, n\}$
and  $[n] \triangleq \{1, \ldots, n\}$. 
%
%

The use of the indexes is consistent through the paper, with the independent variable $u$/$k$/$n$/$t$ iterating over the set $[U]$/$[K]$/$[N]$/$[T]$. Accordingly, we  use the shorthand $\sum_{u \in [U]}x_u \triangleq \sum_{u} x_u$ after the first time the notation is used.

\section{Preliminaries}
\label{sec:Preliminaries}
\subsection{Federated Learning (FL) setting}
\label{sec:FL Setting}

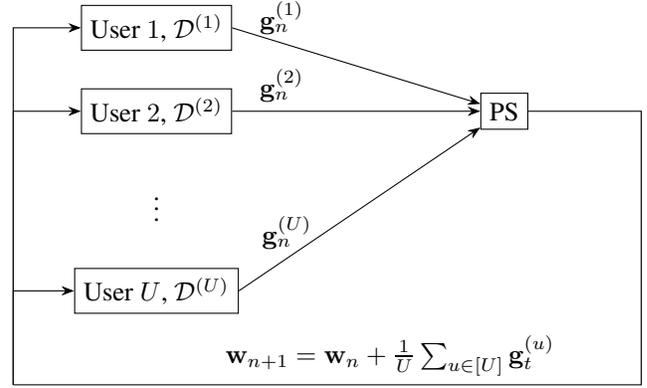
\begin{figure}
\centering

\begin{tikzpicture}
\node[draw, rectangle] (UE1) {User $1$, $\Dcal^{(1)}$};
\node[draw, rectangle, below=0.5cm of UE1] (UE2) {User $2$, $\Dcal^{(2)}$};
\node[below=0.5cm of UE2] (dots) {$\vdots$};
\node[draw, rectangle, below=0.5cm of dots] (UE3) {User $U$, $\Dcal^{(U)}$};
\node[draw, rectangle, right=3.3cm of UE2] (PS) {PS};

\draw[-Stealth] (UE1.east) -- (PS) node[pos=0.2, above] {$\gv_n^{(1)}$};
\draw[-Stealth] (UE2.east) -- (PS) node[pos=0.2, above] {$\gv_n^{(2)}$};
\draw[-Stealth] (UE3.east) -- (PS) node[pos=0.2, above] {$\gv_n^{(U)}$};
\coordinate (branchingPoint) at ($(UE3.east) + (-3,-1.25)$);

\draw[-] (PS.east) -- ++(1.5,0) |- (branchingPoint) node[pos=0.7, above] {$\wv_{n+1}=\wv_n+\f 1 U \sum_{u \in [U]} \gv_t^{(u)}$};

\draw[-Stealth] (branchingPoint) |- (UE1);
\draw[-Stealth] (branchingPoint) |- (UE2);
\draw[-Stealth] (branchingPoint) |- (UE3);

\end{tikzpicture}









\caption{The Federated Learning (FL) setting Sec. \ref{sec:FL Setting}.}
\label{fig::FL Setting}

\end{figure}

Consider the scenario in Fig. \ref{fig::FL Setting} with $U$ remote users, each possessing a local dataset
\ea{
\Dc^{(u)} = \left\{\lb \dv_{j}^{\lb u\rb},v_j^{\lb u\rb}\rb\right\}_{j\in [J]},
}
where $\Dc^{(u)}$  includes
$J$
pairs, each comprising a data point 
$\dv_{j}^{\lb u\rb}$
and the label $v_{j}^{(u)}$
for $u \in [U]$.  
Users collaborate with the PS to minimize the loss function $\Lcal$ as evaluated across all the local datasets and over the choice of the model $\wv \in \Rbb^d$,
that is 
\ea{
\Lcal(\wv) =  \f 1 {U J} \sum_{u } \sum_{j} \Lcal(\wv;\dv^{(u)}_{j},v^{(u)}_{j}).
\label{eq:loss}
}
%
For the \emph{loss function} $\Lcal$ in the LHS of \eqref{eq:loss}, we assume that there exists a unique minimizer $\wv^*$, which we referred to as the \emph{optimal model}.

A common approach for numerically determining this unique minimizer, $\wv^*$, is through the iterative application of (synchronous) SGD.
In the SGD algorithm, the model parameter $\wv$ is updated at each iteration $n$, 
by taking a step toward the negative direction of the stochastic gradient vector,   that is 
\ea{
\wv_{n+1}=\wv_{n}-\eta_n  \gv_n,
\label{eq:SGD}
}
for $n \in [N]$, a choice of initial model $\wv_{0}$, and where  $\gv_n$ is the stochastic gradient of $\Lcal(\cdot)$ evaluated in $\wv_{n}$, that is
$\Ebb\lsb \gv_t\rsb=  \nabla\Lcal\lb\wv_n,\Dcal^{(u)}\rb$.
Finally, $\eta_n$ in \eqref{eq:SGD} is an iteration-dependent step size, the \emph{learning rate}.

In the FL setting, the SGD iterations  are distributed among $U$ users and are orchestrated by PS as follows:
(i) each user $u \in[U]$ receives the current model estimate, $\wv_n$ of the optimal model $\wv^*$ over the infinite capacity link from the PS.
The user $u$ then (ii) accesses its  local dataset $\Dc^{(u)}$ and computes the local stochastic gradient $\gv_n^{(u)}$.
Finally (iii) each node communicates the gradient estimate $\gv_{n}^{(u)}$ to the PS which then computes the term $\gv_n$  as
\ea{
\gv_n=\f{1}{U}\sum_{u} \gv_n^{(u)},
\label{eq:aggregate}
}
and uses $\gv_t$ to update the model estimate as in \eqref{eq:SGD}.
We refer to the above FL training algorithm as  \emph{federate averaging} (FedAvg)  \cite{mcmahan2017communication}.

\subsection{The Random Access CHannel (RACH) setting}
\label{sec:RACH setting}
\begin{figure*}
    \centering
    \begin{tikzpicture}[>=Stealth]
        \draw[thick] (0,0) -- (6,0); 
        \draw[thick] (9,0) -- (15,0); 

        \foreach \x in {0, 1, ..., 6} {
            \draw (\x, 0.1) -- (\x, -0.1);
        }

        \foreach \x in {0, 3, 6} {
            \draw[very thick] (\x, 0.2) -- (\x, -0.2);
        }

        \node at (0, -0.5) {$0$};
        \node at (3, -0.5) {$T$};
        \node at (6, -0.5) {$2T$};

        \node at (7.5, 0) {$\ldots$};

        \foreach \x in {9, 12, 15} { 
            \draw (\x, 0.1) -- (\x, -0.1);
            \draw[very thick] (\x, 0.2) -- (\x, -0.2);
        }
        \node at (9, -0.5) {$(N-2)T$};
        \node at (12, -0.5) {$(N-1)T$};
        \node at (15, -0.5) {$NT$};

        \foreach \x/\w in {0/\wv_0, 3/\wv_1, 6/\wv_2, 9/\wv_{N-2}, 12/\wv_{N-1}, 15/\wv_N} {
            \filldraw[fill=black] (\x,0) circle (1pt);
            \node[below] at (\x, -1) {$\w$}; 
            \draw[->, thick] (\x, -1.1) -- (\x, -0.65); 
        }

        \draw [decorate,decoration={brace,amplitude=10pt,raise=5pt},yshift=0pt] (3,0.2) -- (6,0.2) node [black,midway,yshift=25pt] {\footnotesize 1 time-frame};

        \draw [decorate,decoration={brace,amplitude=10pt,raise=5pt},yshift=0pt] (5.75,-0.2) -- (3.25,-0.2) node [black,midway,yshift=-25pt] {\footnotesize $K$ slots};

    \end{tikzpicture}
    \caption{The RACH setting discussed  in Sec. \ref{sec:RACH setting}.}
    \label{fig:RACH setting}
\end{figure*}
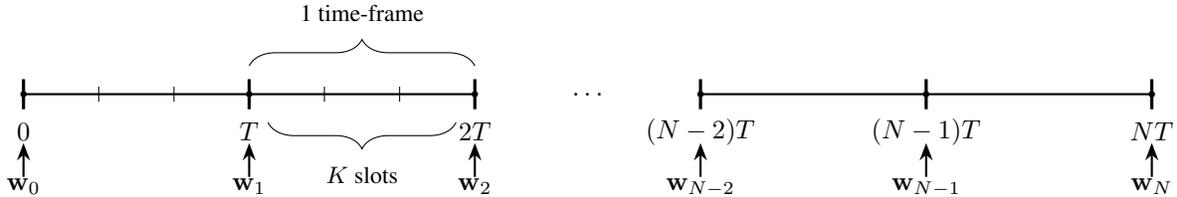

%
Throughout our discussion, we focus on a RACH which implements a (framed) slotted ALOHA policy, as depicted in Fig. \ref{fig:RACH setting}. 
%
%
In particular, we assume the overall time-horizon to be sub-divided in $N$ time-frames, each of duration $T$ (sec), and comprising $K$ slots of equal duration. The transmission parameters of the learning agents are set such that each of them can attempt delivery of a single packet over a slot. Accordingly, given the RACH transmission rate $\Rsf$ floats/sec-- see Rem. \ref{rem:Rate definition}-- we have that $T \Rsf/K$ floats can be transmitted in each slot.

In time-frame $n$ user $u$ transmits in slot $k$ with probability  $p^{(u)}_n$,  iid over $k \in [K]$. 
When more that than one user transmits in the same slot, then a packet collision occurs and transmission fails for all transmitting users.
The protocol is conceptually represented in Fig. \ref{fig:RACH setting} (ignore the $\wv_n$ for the time being).

As discussed, the iid transmission probability in each slot in a time-frame shall be chosen to maximize the throughput, i.e., the average number of packets decoded at the PS per slot. 
If $U_A$ users transmit in a time-frame, that is these users are \emph{active} i.e. $p^{(u)}_n>0$, then slotted ALOHA attains the highest throughput of $\sim 1/e$ (packet/slot) for $p_{n}^{(U)}=1/{U_A}$ \cite{Abramson77_PacketBroadcasting}.

\begin{remark}[\bf Rate definition]
\label{rem:Rate definition}
%
We assume that the rate of the RACH is measured in floats/second.
%
This is done to simplify the presentation and avoid specifying the number of bits used to represent each unit of information.
\end{remark}

\section{RACH-FL Model}
\label{sec:RACH-FL Model}

In this paper, we study the scenario in which the model updates for the FL setting in Sec. \ref{sec:FL Setting} are transmitted over the RACH channel in Sec. \ref{sec:RACH setting}. 
More precisely, $N$ rounds of FedAvg are performed in the $N$ time-frames, as in Fig. \ref{fig:RACH setting}.
The remote users transmit the model updates to the PS using slotted ALOHA in the $K$ slots comprising each time-frame.
The time-frame $n \in [N]$ begins at wall-clock time $t=nT$.
At time instant $t=0$, the model initialization $\wv_0$ is transmitted from the PS to the remote users.
At the beginning of the time-frame $n$, the model update $\wv_n$ is transmitted from the PS to all the remote users as indicated in Fig. \ref{fig:RACH setting}. 
Along with the model updates, the PS can also transmit a set of parameters $\hv$ to control the user policy. 
During the time-frame, the $U$ users employ a certain policy to (i) compress their local gradients to to meet the rate constraint over the ALOHA slot, (ii) determine the transmission probability over the slot, and (iii) address the error introduced by both the gradient compression and the channel contention.  
This policy can depend on the parameters $\hv$ transmitted from the PS to all the users. 

    

A few remarks are in order:

\begin{itemize} [leftmargin=*]

\item
{[\bf Coordination and scheduling:]} 
In the RACH-FL problem formulation, the training process features scalable, dynamic, and heterogeneous client participation. This environment renders user coordination, scheduling, and transmission feedback impractical.
Consequently, the protocol design is constrained to having the PS send a set of hyperparameters along with the model updates. 
These hyperparameters then dictate user behavior for a specified time-frame according to a pre-determined policy.

\item
{[\bf Model parameters:]} here we assume that the parameters  $\Rsf,T,U,K,N$ are fixed and cannot be changed by the learning architect. 
This is because we envision a system in which the medium is used to support a number of learning instances -- possibly also in parallel.   

\item {[\bf iid transmission probability:]} users are limited to iid transmission probabilities in each time-frame. 
That is, user $u$ in time-frame $n$ transmits with iid probability $p_n^{(u)}$ for each slot $k \in [K]$.
This policy maximizes throughput under the assumption that users do not cooperate and that the PS does not schedule users into certain slots.

 \item {[\bf Computation time]}
For simplicity, we assume that all computation times, e.g. gradient updates, gradient compression, and model aggregation, are negligible.
%
%
%
\end{itemize}
%

The RACH-FL model is defined by the set of parameters  $\Rsf,T,U,K,N$, along with a learning problem as in \eqref{eq:loss}. 
Given a RACH-FL model, the learning architect is tasked with designing a policy for (i) compressing the gradient, (ii)  determining the transmission probability, and (iii) devising an error feedback mechanism to minimize the loss in \eqref{eq:loss}.
The design of such policies will be tackled in Sec. \ref{sec:Age-of-Gradient Policy}.



%

\medskip

A summary of the quantities for the RACH-FL model is provided in Table \ref{table:summary symbols}.

\begin{table}[h!]
\centering
\begin{tabular}{|l|c|}
\hline
\textbf{Quantity} & \textbf{Description}  \\
\hline
Num. of users & $U$ \\
\hline
Num. of time-frames/FL iterations & $N$ \\
\hline
Num. of ALOHA slots per time-frame & $K$ \\
\hline
RACH rate  & $\Rsf$ \\
\hline
Length of a time-frame & $T$ \\
\hline
Local dataset size & $J$ \\
\hline
\end{tabular}
\caption{A summary of the parameters of the RACH-FL model introduced in  Sec. \ref{sec:RACH-FL Model}.}
\label{table:summary symbols}
\end{table}

\subsection{Introductory Example}
\label{sec:Introductory Example}

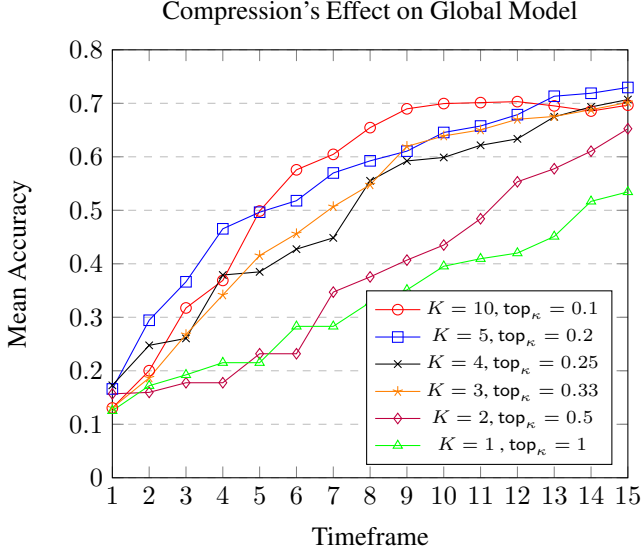
\begin{figure}
    \centering
\begin{tikzpicture}
\begin{axis}[
    title={Compression's Effect on Global Model},
    xlabel={Timeframe},
    ylabel={Mean Accuracy},
    xmin=1, xmax=15,
    ymin=0, ymax=0.8,
    xtick={1,2,3,4,5,6,7,8,9,10,11,12,13,14,15},
    ytick={0,0.1,0.2,0.3,0.4,0.5,0.6,0.7,0.8},
    legend pos=south east,
    legend style={font=\scriptsize},
    ymajorgrids=true,
    grid style=dashed,
]

\addplot[
    color=red,
    mark=o,
    ]
    coordinates {
    (1,0.13046)(2,0.20015999)(3,0.31744)(4,0.36886)(5,0.49866)(6,0.57566)(7,0.60456001)(8,0.6545)(9,0.6897)(10,0.6995)(11,0.7013)(12,0.70316001)(13,0.69516)(14,0.68548)(15,0.69652001)
    };
    \addlegendentry{$K=10$, $\topK=0.1$}

\addplot[
    color=blue,
    mark=square,
    ]
    coordinates {
    (1,0.16602)(2,0.29444)(3,0.36634)(4,0.4654)(5,0.49644)(6,0.51789999)(7,0.56975999)(8,0.59243999)(9,0.6106)(10,0.64545999)(11,0.65754)(12,0.67874001)(13,0.71322001)(14,0.71880001)(15,0.72958001)
    };
    \addlegendentry{$K=5$, $\topK=0.2$}
\addplot[
    color=black,
    mark=x,
    ]
    coordinates {
    (1,0.1728)(2,0.24752)(3,0.26052)(4,0.379)(5,0.3849)(6,0.42736)(7,0.44848)(8,0.55508001)(9,0.59226)(10,0.5988)(11,0.6215)(12,0.63354001)(13,0.67472)(14,0.69324001)(15,0.70700001)
    };
    \addlegendentry{$K=4$, $\topK=0.25$}
    
\addplot[
    color=orange,
    mark=star,
    ]
    coordinates {
    (1,0.1308)(2,0.18662)(3,0.26774)(4,0.34182)(5,0.41543999)(6,0.45628)(7,0.50688001)(8,0.54769999)(9,0.61948)(10,0.639)(11,0.65026)(12,0.67016)(13,0.67542)(14,0.68801999)(15,0.70165999)
    };
    \addlegendentry{$K=3$, $\topK=0.33$}
\addplot[
    color=purple,
    mark=diamond,
    ]
    coordinates {
    (1,0.15688)(2,0.1596)(3,0.17762)(4,0.17762)(5,0.23182)(6,0.23182)(7,0.34706)(8,0.37546)(9,0.40688)(10,0.43503999)(11,0.48417998)(12,0.55323999)(13,0.57763999)(14,0.61043999)(15,0.65244)
    };
    \addlegendentry{$K=2$, $\topK=0.5$}
    
\addplot[
    color=green,
    mark=triangle,
    ]
    coordinates {
    (1,0.1252)(2,0.172)(3,0.19236)(4,0.21512)(5,0.21512)(6,0.28304)(7,0.28304)(8,0.32798)(9,0.35116)(10,0.3954)(11,0.40948)(12,0.42004)(13,0.45105999)(14,0.51672)(15,0.53434001)
    };
    \addlegendentry{$K=1$ , $\topK=1$}

\end{axis}
\end{tikzpicture}
    \caption{The RACH-FL example discussed in Sec. \ref{sec:Introductory Example} in which $10$ users transmit their model updates over the RACH using a transmission probability of $1/10$. 
    Each curve represents a different number of slots $K$ per time-frame. For each $K$, the model updates are sparsified so as to keep the channel rate the same. 
    The simulation settings are described in  Sec. \ref{sec:DNN Setting}.
    }
    \label{fig:various k}
\end{figure}

Before delving further in the design of joint training/communication strategies for the RACH-FL setting, let us provide a rather intuitive example on the relevance of the RACH-FL model.
Consider the training process as in Sec.~\ref{sec:FL Setting}, and assume that the user accesses the RACH using a slotted ALOHA protocol in which all users transmit in slot with probability $p_{t}^{(u)}=p=1/U$. 
As discussed in Sec. \ref{sec:RACH setting}, this a reasonable choice of the transmission probability as it maximizes the fraction of slots containing successful transmissions, i.e. the throughput. 
In order to address the rate constraint, $\topK$ sparsification is applied so that
\ea{
{|\topK(\gv_n^{(u)})|}= \f {\Rsf T} K,
}
meaning that the size of the sparsified model update equals the amount of information that can be transmitted in each of the $K$ slots in a time-frame. 
%
%
Arguably, the transmission strategy described above is a meaningful benchmark for the RACH-FL setting in which the gradient compression strategy and the random access protocol are both efficient and fully scalable.
Notice that the two policies, i.e. training and communication, operate independently: as such this represent a baseline performance for the development of more competitive strategies.
With the above setting in mind, we can discuss the RACH-FL model performance for varying $K$.
By varying the value of $K$ --and adjusting the sparsification level accordingly-- one can observe the resulting change in the training accuracy as a function the time-frame index.
This numerical experiment is depicted in Fig. \ref{fig:various k} -- the simulation settings are presented later in Sec. \ref{sec:DNN Setting}.
From Fig. \ref{fig:various k} we gather some interesting insights.
First, we observe that accuracy increases with $K$.
This is very much expected as DNNs models are generally highly parametrized and $\topK$ sparsification generally improves performance for small values of $K$.
Secondly, the performance for large values of $K$ shows some interesting trade-off, with $K=5$ slots providing better performance at the beginning of training ($N \leq 5$), while $K=10$ yielding better results in the latter part of training ($5< N < 12$). 
%
%

We believe that this example highlights the interesting tension arising from channel contention and training accuracy. 
In the next section we shall focus on the design of a joint training/communication strategy that maximizes the learning performance in the RACH-FL model.

\medskip

\section{Age of Gradient (AoG) Strategy}
\label{sec:Age-of-Gradient Policy}


In this section, we introduce the core contribution of the paper: the Age-of-Gradient (AoG) strategy. Inspired by the Age of Information (AoI) literature \cite{yates2021age}, which focuses on optimizing the timeliness of status updates in constrained systems \cite{kaul2012real}, AoG adapts this concept to RACH-FL setting. 
AoI has provided valuable insights into optimizing update freshness across various systems, including server systems, queueing networks, and multi-hop wireless networks. 
For RACH-FL, we define "freshness of the gradient" as a combination of the current gradient and the errors due to gradient compression and user selection. Detailed insights into the AoG strategy are provided in this section.

The AoG strategy is defined as follows:

\begin{itemize} [leftmargin=*]

\item {[\bf Gradient compression strategy]} the gradient entries are sparsified to $\Rsf T/K$ floats to fit the slot of the slotted ALOHA protocol using $\topK$ sparsification over the memory (see next) plus local gradient term \cite{stich2018sparsified}. Specifically, we have
\ea{
\ghv_n^{(u)} = {\rm top}_{\Rsf T/K} \lb \gv_n^{(u)}+ \mv_n^{(u)}\rb,
\label{eq:compress AoG}
}
where $\topK(\xv)$ extract the $\kappa$  elements with the largest magnitude from $\xv$. 
Accordingly, the PS only accumulates the compressed gradients  from the active users (see next), FedAvg at the PS  is modified in \eqref{eq:SGD}
as 
\begin{equation}
    \whv_{n+1} = \whv_{n} - \eta_{n} \sum_{k\in \Kcal_n}\hat{\mathrm{\gv}}_n^{(u)},\label{eq:aggregation compression}
\end{equation}
where $\Kcal_n$ is the set of active users at round $n$. 

\item   {[\bf Error correction strategy]} 
To address the difference between $\wv_t$ in \eqref{eq:SGD} and $\whv_t$ in \eqref{eq:aggregation compression}, we use an error feedback mechanism at the remote user \cite{stich2020error}. 
We define $\mv_t^{u}$ in \eqref{eq:aggregation compression} as
\ea{
\mv_{n+1}^{(u)} = \gamma \mv_n^{(u)} +   \ev_{n}^{(u)},
\label{eq:memory accumulation AoG}
}
where
\ea{
\ev_n^{(u)} = \lcb \p{
\gv_n^{(u)}-\ghv_n^{(u)}  &  u \in \Kcal_n \\
\gv_n^{(u)}  & u \not \in \Kcal_n,
}\rnone
}
to correct the error between $\wv_n$ and $\whv_n$.
The term $\ev_n^{(u)}$ corresponds to the error arising in FedAvg from (i) gradient compression,  if $u \in \Kcal_n$, or (ii) user exclusion, if $u \not \in \Kcal_n$.  
We refer to $\gamma$ as the \emph{forget coefficient}. 
\item  {[\bf User selection strategy]}
This strategy is the core of the AoG policy.
Together with the model $\wv_n$, the PS also sends a threshold  $\tau_{n+1}$ and a transmission probability $\rho_{n+1}$.
%
A user the becomes active only if
\ea{
\| \gv_n^{(u)} +\mv_n^{(u)}\|_2 \geq \tau_n.
\label{eq:threshold policy}
}
In this case, the user will contend over the $n$ time-frame, attempting transmission of its sparsified gradients in each slot of the frame. 
We indicate the set of active users as $\Kcal_n$, and denote by  $\tau_n$ and $\rho_n$  the \emph{activation threshold} and \emph{active user transmission probability}, respectively.
\item {\bf [Fallback Protocol:]} in the unlikely event that  all slots over the RACH result in a collision for time-frame $n$, the server will send  $\wv_n=\wv_{n-1}$. In this case we can assume that the same user selection strategy is again employed. 
\end{itemize}

Two final remarks are in order. First, we note that \eqref{eq:threshold policy} uses the magnitude of the current local gradient plus the memory as a measure of the freshness of the contribution of the user. Second, we recall from Sec. \ref{sec:RACH setting} that the parameter $\rho_n$ should be chosen as the inverse of the number of active users, $|\Kcal_n|$. 
In the next section, we discuss the optimization of the activation threshold and the active user transmission probability.

\subsection{User Selection strategy Optimization}
\label{sec:Transmission Probability Optimization}
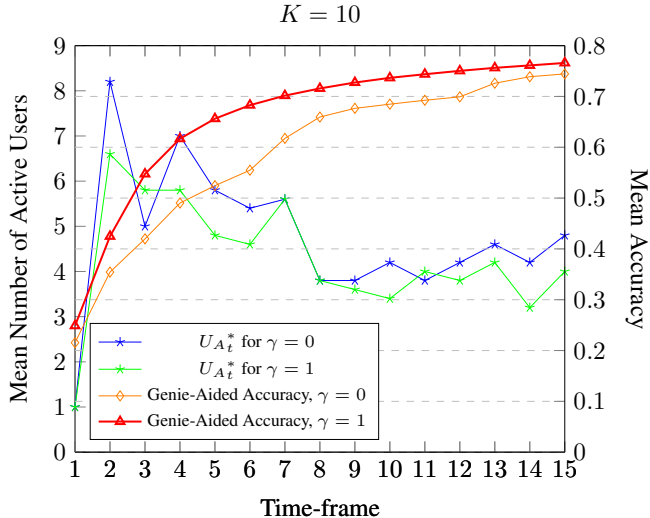
\begin{figure}
    \centering
    \begin{tikzpicture}[scale=0.95]
    \begin{axis}[
    title={$K = 10$},
        axis y line*=left,
        xmin=1, xmax=15,
        ymin=0, ymax=9,
        xtick={1,2,3,4,5,6,7,8,9,10,11,12,13,14,15},
        ytick={0,1,2,3,4,5,6,7,8,9},
        xlabel=Time-frame,
        ylabel=Mean Number of Active Users,
          y label style={at={(axis description cs:0.1,.5)},anchor=south},
    ]
    \addplot[color=blue, mark=star] coordinates {
        (1, 1.0)
        (2, 8.2)
        (3, 5.0)
        (4, 7.0)
        (5, 5.8)
        (6, 5.4)
        (7, 5.6)
        (8, 3.8)
        (9, 3.8)
        (10, 4.2)
        (11, 3.8)
        (12, 4.2)
        (13, 4.6)
        (14, 4.2)
        (15, 4.8)
    };
    \label{memoryless}

    \addplot[color=green, mark=star] coordinates {
        (1, 1)
        (2, 6.6)
        (3, 5.8)
        (4, 5.8)
        (5, 4.8)
        (6, 4.6)
        (7, 5.6)
        (8, 3.8)
        (9, 3.6)
        (10,3.4)
        (11,4.0)
        (12,3.8)
        (13,4.2)
        (14,3.2)
        (15,4.0)
    };
    \label{gamma1}
    \end{axis}
    
    \pgfplotsset{every axis y label/.append style={rotate=180,yshift=9.5cm}}
    \begin{axis}[
        axis y line*=right,
        xmin=1, xmax=15,
        ymin=0, ymax=0.8,
        xtick={1,2,3,4,5,6,7,8,9,10,11,12,13,14,15},
        ytick={0,0.1,0.2,0.3,0.4,0.5,0.6,0.7,0.8},
        xlabel=Time-frame,
        ylabel=Mean Accuracy,
        legend pos=south west,
        legend style={font=\scriptsize},
        ymajorgrids=true,
        grid style=dashed,
          y label style={at={(axis description cs:-0.1,.5)},anchor=south},
    ]
    \addlegendimage{/pgfplots/refstyle=memoryless}\addlegendentry{${U_A}_t^*$ for $\gamma=0$}
    \addlegendimage{/pgfplots/refstyle=gamma1}\addlegendentry{${U_A}_t^*$ for $\gamma = 1$}
    \addplot[color=orange, mark=diamond] coordinates {
        (1, 0.21543734)
        (2, 0.35440512)
        (3, 0.41964904)
        (4, 0.49021966)
        (5, 0.52444324)
        (6, 0.55481194)
        (7, 0.61745004)
        (8, 0.6595054)
        (9, 0.67657006)
        (10, 0.68478636)
        (11, 0.6924799)
        (12, 0.69918556)
        (13, 0.72599702)
        (14, 0.7387599)
        (15, 0.74435536)
    };
    \addlegendentry{Genie-Aided Accuracy, $\gamma=0$}

    \addplot[
        color=red,
        mark=triangle,
        thick
        ]
        coordinates {
        (1,0.248736)
        (2,0.424724)
        (3,0.547344)
        (4,0.616776)
        (5,0.656220)
        (6,0.682800)
        (7,0.701592)
        (8,0.715752)
        (9,0.727348)
        (10,0.736572)
        (11,0.743692)
        (12,0.750360)
        (13,0.756260)
        (14,0.761064)
        (15,0.765912)
        };
        \addlegendentry{Genie-Aided Accuracy, $\gamma=1$}
    \end{axis} 
    \end{tikzpicture}
    \caption{
    The performance of the genie-aided policy as discussed in Sec. \ref{sec:Transmission Probability Optimization}.
    The simulation setting are as in Sec. \ref{sec:DNN Setting}.}
    \label{fig: active users vs accuracy}
\end{figure}

The crucial insight of the paper lies in the choice of which users shall be active over the RACH in each time-frame.
We begin by numerically investigating this non-trivial problem through a genie-aided policy.
Assume that the PS collects all the gradient-plus-memory magnitudes from the users, selects the active users, simulates the channel multiple times, and observes which choice of the number of active users yields the best accuracy when averaged over the RACH behavior. 
This policy yields $\Kcal_t^*$, the optimal set of active users.
From $\Kcal_t^*$, one immediately obtains the optimal choice for activation threshold  and active user transmission probability, $\rho^*$ is obtained as  
\ea{
\rho_t^* = \f 1 {| {U_A}_t^* | },
}
for ${U_A}_t^*=| \Kcal_t^* |$.
%
%
The results of this numerical procedure are depicted in  Fig. \ref{fig: active users vs accuracy}.
We refer to this approach as the \emph{genie-aided} policy. 
In Fig. \ref{fig: active users vs accuracy}  we also plot the performance with forget coefficient $\gamma=1$. Notably, the behavior remains similar, although memory allows for a generally lower number of active users.

\begin{figure}

\begin{tikzpicture}[scale=0.95]     
\begin{axis}[
    title={$K = 10$},
    xlabel={Time-frame},
    ylabel={Mean Accuracy},
    xmin=1, xmax=15,
    ymin=0.1, ymax=0.8,
    xtick={1,2,3,4,5,6,7,8,9,10,11,12,13,14,15},
    ytick={0.1,0.2,0.3,0.4,0.5,0.6,0.7,0.8},
    legend pos=south east,
    legend style={font=\scriptsize},
    ymajorgrids=true,
    grid style=dashed,
]

\addplot[
    color=red,
    mark=triangle,
    ]
    coordinates {
        (1,0.15423)
        (2,0.3861064)
        (3,0.4542140)
        (4,0.4942888)
        (5,0.5925308)
        (6,0.6247340)
        (7,0.6757796)
        (8,0.6831256)
        (9,0.7135192)
        (10,0.7245312)
        (11,0.7412420)
        (12,0.7499964)
        (13,0.7568760)
        (14,0.7615380)
        (15,0.7669828)
        };
    \addlegendentry{fix/random $U_A=10$, $\gamma = 1$}

\addplot[
    color=green,
    mark=o,
    ]
    coordinates {
    (1,0.2165924)
        (2,0.378818)
        (3,0.4592748)
        (4,0.5247492)
        (5,0.5943688)
        (6,0.6391516)
        (7,0.680832)
        (8,0.6904988)
        (9,0.7154196)
        (10,0.7269588)
        (11,0.7458832)
        (12,0.7514816)
        (13,0.7596764)
        (14,0.7633252)
        (15,0.7684416)
    };
    \addlegendentry{fix/random $U_A=5$, $\gamma = 1$}

\addplot[
    color=brown, mark=triangle,
    ]
    coordinates {
    (1,0.13046)(2,0.20015999)(3,0.31744)(4,0.36886)(5,0.49866)(6,0.57566)(7,0.60456001)(8,0.6545)(9,0.6897)(10,0.6995)(11,0.7013)(12,0.70316001)(13,0.69516)(14,0.68548)(15,0.69652001)
    };
    \addlegendentry{fix/random $U_A=10$, $\gamma=0$}

\addplot[
        color=blue, mark=o,
        ]
        coordinates {
        (1,0.2433)(2,0.3602)(3,0.3882)(4,0.4500)(5,0.4897)(6,0.5288)(7,0.5435)(8,0.5643)(9,0.6134)(10,0.6136)(11,0.6705)(12,0.6712)(13,0.6945)(14,0.7241)(15,0.7212)
        };
    \addlegendentry{fix/random $U_A=5$, $\gamma=0$}

\addplot[
            color=blue,
        mark=*,
        mark options={scale=0.75},
    ]
    coordinates {
        (1,0.203096)
            (2,0.392196)
            (3,0.523512)
            (4,0.601912)
            (5,0.645108)
            (6,0.674484)
            (7,0.694136)
            (8,0.70918)
            (9,0.7212)
            (10,0.731032)
            (11,0.73938)
            (12,0.746336)
            (13,0.752236)
            (14,0.757824)
            (15,0.762876)
    };
    \addlegendentry{AoG Policy $\gamma=1$}

    \addplot[
        color=pink,
        mark=star,
        thick
        ]
        coordinates {
        (1,0.248736)
        (2,0.424724)
        (3,0.547344)
        (4,0.616776)
        (5,0.656220)
        (6,0.682800)
        (7,0.701592)
        (8,0.715752)
        (9,0.727348)
        (10,0.736572)
        (11,0.743692)
        (12,0.750360)
        (13,0.756260)
        (14,0.761064)
        (15,0.765912)
        };
        \addlegendentry{Genie-Aided $\gamma=1$}
 \end{axis}
\end{tikzpicture}
\caption{The performance of the AoI policy, as compared with the baselines discussed in Sec. \ref{sec:baselines} for $K=10$. The simulation settings are as in Sec. \ref{sec:DNN Setting}. }
\label{fig:threshold policy}

\end{figure}
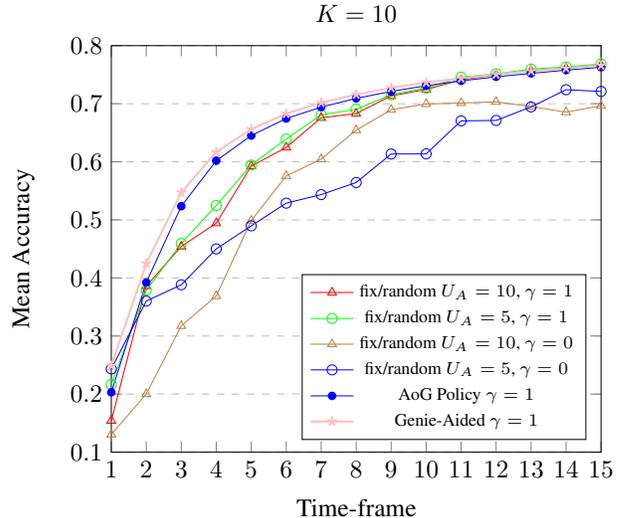

Once we have obtained a numerical description of $\Kcal_t^*$, also averaged across various network initialization and channel simulations, we  can then derive some insight on the relationship between $\Kcal_t^*$ and choose of $\tau_n$ and $\rho_n$.
As we shall argue in Sec.~\ref{sec:Numerical Results}, a reasonable choice for this policy is to set 
\ea{
\Uh_t =  c \| \gv_t|_2 + U_{\min},
\label{eq:U hat}
}
where $\Uh_t$ is meant to approximate $\| \Kcal_t^* \|$. The values of $\tau_t$ can then be selected to activate the $\Uh_t$ users with the largest magnitude of gradient plus memory by monitoring the variance of the gradient updates $\gv_{t-1}^{(u)}$ received at the PS.  
Furthermore, $\rho_t$ is set to $1/\Uh_t$.
We remark that the simple optimization of the threshold policy proposed in \eqref{eq:U hat} is able to well emulate the genie-aided behavior: this is depicted in Fig. \ref{fig:threshold policy}. The simulation settings are discussed in Sec. \ref{sec:DNN Setting}, the baselines are discussed in Sec. \ref{sec:baselines}

\section{Numerical Results}
\label{sec:Numerical Results}

In this section, we expand the numerical evaluation in Sec.~\ref{sec:Age-of-Gradient Policy} to different settings, deriving further insights on the potential of the proposed AoG policy.

\subsection{Simulation Settings}
\label{sec:DNN Setting}
%
%
All simulations in the paper are performed for the DNN presented next. 

\begin{itemize} [leftmargin=*]

\item {\bf [DNN setting]}
We consider the distributed training of the VGG16 network \cite{simonyan2014very}, comprised of $16$ layers: $13$ convolutional layers followed by $3$ fully connected layers. 
The total number of parameters is $1.38 \times 10^8$.   
The network is trained over the CIFAR-10 for image classification \cite{cifar10}. 
%
%
The network is trained using the Adam optimizer with learning rate $\eta=0.0001$.
Training is repeated with 5 random seeds. 
The number of simulation of the RACH vary between 5 and 100.
%

\item {\bf [FL setting]}
We consider 10 remote users. The local dataset 
have equal size -- $3000$ -- obtained by sampling without replacement from the original CIFAR-10 dataset, which has  $30000$ samples. 

\item {\bf [RACH setting]} We consider the scenario in which the whole set of VGG16 parameters, i.e. $1.38 \times 10^8$ floats, can be transmitted over a whole time-frame. 
For this reason, the capacity of every slot is $\Rsf =1.38 \times 10^8/T$.
\end{itemize}
Note that one could normalize $T$ to be unitary and adjust the rate $\Rsf$ accordingly. 
Due to space limitations, we do not discuss this normalization here.

\subsection{Baselines}
\label{sec:baselines}


\begin{itemize} [leftmargin=*]

\item {\bf [genie-aided]}: the active users are chosen to be $\Kcal_t^*$ as determined in Sec. \ref{sec:Transmission Probability Optimization}. 

\item {\bf [fix/random -- $U_A=m$]}: $m$ active users are randomly selected at each time-frame for transmission.
This protocol can be implemented in a distributed manner using a random number generator and has no further complexity or memory requirement.%

\item {\bf [$\gamma=0$ vs $\gamma=1$]}: in Appendix B of the supplementary material we show that the choice $\gamma=1$ in \eqref{eq:memory accumulation AoG} maximizes the learning performance in the setting of Sec. \ref{sec:DNN Setting}.
Nonetheless, assuming that the users have access to a local memory is unrealistic in some applications. Accordingly we also simulate the case $\gamma=0$.

\end{itemize}






\clearpage

\subsection{AoG performance comparison}

Fig. \ref{fig:threshold policy k=5} is the analog of Fig. \ref{fig:threshold policy} but for $K=5$, instead of $K=10$.
The $K=5$ case has more informative gradients than $K=10$ and the joint optimization of the training/communication strategy allows for larger improvements over the baselines.

\begin{figure}

\begin{tikzpicture} 
\begin{axis}[
    title={$K = 5$},
    xlabel={Time-frame},
    ylabel={Mean Accuracy},
    xmin=1, xmax=15,
    ymin=0.1, ymax=0.85,
    xtick={1,2,3,4,5,6,7,8,9,10,11,12,13,14,15},
    ytick={0.1,0.2,0.3,0.4,0.5,0.6,0.7,0.8},
    legend pos=south east,
    legend style={font=\scriptsize},
    ymajorgrids=true,
    grid style=dashed,
]

\addplot[
    color=red,
    mark=triangle,
    ]
    coordinates {
        (1,0.1359)
        (2,0.3643)
        (3,0.4939)
        (4,0.5456)
        (5,0.5733)
        (6,0.6249)
        (7,0.6444)
        (8,0.6717)
        (9,0.6727)
        (10,0.7112)
        (11,0.7262)
        (12,0.7371)
        (13,0.7379)
        (14,0.7589)
        (15,0.7723)
        };
    \addlegendentry{fix/random $U_A=10$, $\gamma = 1$}

\addplot[
    color=green,
    mark=o,
    ]
    coordinates {
    (1,0.1945)
            (2,0.3865)
            (3,0.5135)
            (4,0.5694)
            (5,0.5873)
            (6,0.6370)
            (7,0.6543)
            (8,0.6814)
            (9,0.6798)
            (10,0.7168)
            (11,0.7289)
            (12,0.7396)
            (13,0.7403)
            (14,0.7606)
            (15,0.7745)
    };
    \addlegendentry{fix/random $U_A=5$, $\gamma = 1$}

\addplot[
    color=brown, mark=triangle,
    ]
    coordinates {
    (1,0.16602)(2,0.29444)(3,0.36634)(4,0.4654)(5,0.49644)(6,0.51789999)(7,0.56975999)(8,0.59243999)(9,0.6106)(10,0.64545999)(11,0.65754)(12,0.67874001)(13,0.71322001)(14,0.71880001)(15,0.72958001)
    };
    \addlegendentry{fix/random $U_A=10$, $\gamma=0$}

\addplot[
        color=blue, mark=o,
        ]
        coordinates {
            (1,0.1676)
            (2,0.2322)
            (3,0.2561)
            (4,0.2308)
            (5,0.2152)
            (6,0.2971)
            (7,0.3644)
            (8,0.4104)
            (9,0.5361)
            (10,0.5829)
            (11,0.6117)
            (12,0.6280)
            (13,0.6278)
            (14,0.6402)
            (15,0.6510)
        };
    \addlegendentry{fix/random $U_A=5$, $\gamma=0$}

\addplot[
            color=blue,
        mark=*,
        mark options={scale=0.75},
    ]
    coordinates {
        (1,0.2575)
            (2,0.5069)
            (3,0.6115)
            (4,0.6715)
            (5,0.7046)
            (6,0.7280)
            (7,0.7437)
            (8,0.7559)
            (9,0.7652)
            (10,0.7730)
            (11,0.7794)
            (12,0.7854)
            (13,0.7902)
            (14,0.7949)
            (15,0.7989)
    };
    \addlegendentry{AoG Policy $\gamma=1$}

    \addplot[
        color=pink,
        mark=star,
        thick
        ]
        coordinates {
        (1,0.344236)
            (2,0.540216)
            (3,0.637048)
            (4,0.68778)
            (5,0.7171)
            (6,0.736684)
            (7,0.751132)
            (8,0.761756)
            (9,0.770292)
            (10,0.777244)
            (11,0.783724)
            (12,0.788796)
            (13,0.793624)
            (14,0.797928)
            (15,0.801668)
        };
        \addlegendentry{Genie-Aided $\gamma=1$}
 \end{axis}
\end{tikzpicture}
\caption{The performance of the AoI policy, as compared with the baselines discussed in Sec. \ref{sec:baselines} for $K=5$. The simulation settings are as in Sec. \ref{sec:DNN Setting}. }
\label{fig:threshold policy k=5}

\end{figure}

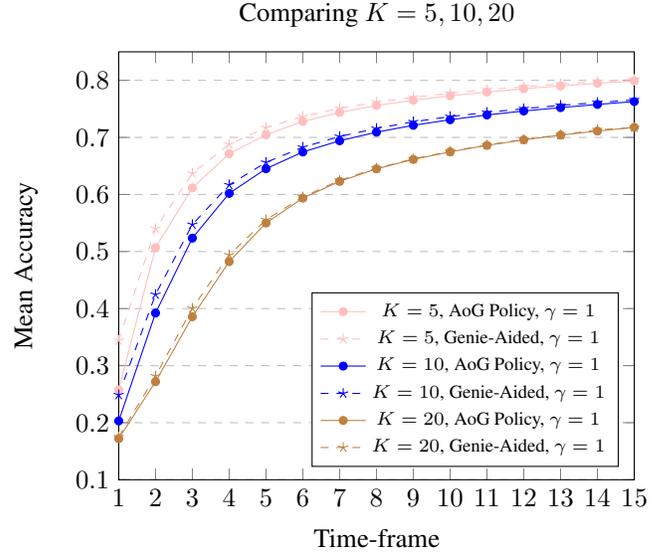
\begin{figure}
\begin{tikzpicture}     
\begin{axis}[
    title={Comparing $K=5,10,20$},
    xlabel={Time-frame},
    ylabel={Mean Accuracy},
    xmin=1, xmax=15,
    ymin=0.1, ymax=0.85,
    xtick={1,2,3,4,5,6,7,8,9,10,11,12,13,14,15},
    ytick={0.1,0.2,0.3,0.4,0.5,0.6,0.7,0.8},
    legend pos=south east,
    legend style={font=\scriptsize},
    ymajorgrids=true,
    grid style=dashed,
]
\addplot[
            color=pink,
        mark=*,
        mark options={scale=0.75},
    ]
    coordinates {
        (1,0.2575)
            (2,0.5069)
            (3,0.6115)
            (4,0.6715)
            (5,0.7046)
            (6,0.7280)
            (7,0.7437)
            (8,0.7559)
            (9,0.7652)
            (10,0.7730)
            (11,0.7794)
            (12,0.7854)
            (13,0.7902)
            (14,0.7949)
            (15,0.7989)
    };
    \addlegendentry{$K=5$, AoG Policy, $\gamma=1$}

    \addplot[
        color=pink,
        mark=star,
        dashed
        ]
        coordinates {
        (1,0.344236)
            (2,0.540216)
            (3,0.637048)
            (4,0.68778)
            (5,0.7171)
            (6,0.736684)
            (7,0.751132)
            (8,0.761756)
            (9,0.770292)
            (10,0.777244)
            (11,0.783724)
            (12,0.788796)
            (13,0.793624)
            (14,0.797928)
            (15,0.801668)
        };
        \addlegendentry{$K=5$, Genie-Aided, $\gamma=1$}
\addplot[
            color=blue,
        mark=*,
        mark options={scale=0.75},
    ]
    coordinates {
        (1,0.203096)
            (2,0.392196)
            (3,0.523512)
            (4,0.601912)
            (5,0.645108)
            (6,0.674484)
            (7,0.694136)
            (8,0.70918)
            (9,0.7212)
            (10,0.731032)
            (11,0.73938)
            (12,0.746336)
            (13,0.752236)
            (14,0.757824)
            (15,0.762876)
    };
    \addlegendentry{$K=10$, AoG Policy, $\gamma=1$}

    \addplot[
        color=blue,
        mark=star,
        dashed
        ]
        coordinates {
        (1,0.248736)
        (2,0.424724)
        (3,0.547344)
        (4,0.616776)
        (5,0.656220)
        (6,0.682800)
        (7,0.701592)
        (8,0.715752)
        (9,0.727348)
        (10,0.736572)
        (11,0.743692)
        (12,0.750360)
        (13,0.756260)
        (14,0.761064)
        (15,0.765912)
        };
        \addlegendentry{$K=10$, Genie-Aided, $\gamma=1$}
\addplot[
            color=brown,
        mark=*,
        mark options={scale=0.75},
    ]
    coordinates {
            (1,0.17204)
            (2,0.271896)
            (3,0.38574)
            (4,0.482352)
            (5,0.549976)
            (6,0.593572)
            (7,0.622956)
            (8,0.644912)
            (9,0.66142)
            (10,0.674712)
            (11,0.686052)
            (12,0.695736)
            (13,0.704076)
            (14,0.7111)
            (15,0.717712)
    };
    \addlegendentry{$K=20$, AoG Policy, 
 $\gamma=1$}

    \addplot[
        color=brown,
        mark=star,
        dashed
        ]
        coordinates {
        (1,0.175016)
            (2,0.28178)
            (3,0.40052)
            (4,0.49394)
            (5,0.555612)
            (6,0.595572)
            (7,0.624732)
            (8,0.646068)
            (9,0.662316)
            (10,0.675476)
            (11,0.686768)
            (12,0.696336)
            (13,0.704284)
            (14,0.71336)
            (15,0.717096)
        };
        \addlegendentry{$K=20$, Genie-Aided,  $\gamma=1$}
 \end{axis}
\end{tikzpicture}

\caption{The performance of the AoI policy, as compared with the genie-aided policy across $K=5,10$ and $20$.
 The simulation settings are as in Sec. \ref{sec:DNN Setting}. }
\label{fig:threshold policy compare}
\end{figure}
In Fig. \ref{fig:threshold policy compare} we compare the AoG policy and the genie-aided policy  for three values of $K$.
Observe how the optimization of the joint training/communication policy improves over the separate strategy discussed in Sec. \ref{sec:Introductory Example} for $K=5$ and $K=10$.

\subsection{Gradient magnitude}

As a final numerical observation, we plot in Fig. \ref{fig:localgrad_mean} the average value of the local-memory plus gradient across various values of $K$.
We note that the progress of the decrease of the weight magnitude is rather well-behaved and can thus be used to infer the parameters of the AoI policy.
This is particularly useful when the number of participating devices varies during the learning process.








\section{Conclusion}

In this paper, we have introduced the problem in which Federated Learning takes place over a Random Access Channels (RACH).
We have term this model as the RACH-FL model and highlighted how this model indeed brings to light trade-offs between the training efficiency and user contention in accessing the communication medium. 
For the RACH-FL model, we introduced the Age-of-Gradient (AoG) strategy, inspired by the concept of Age of Information (AoI), to optimize the communication between remote users and the parameter server (PS).


\begin{figure}[H]
    \centering
    \begin{tikzpicture}[scale=0.95]
\begin{axis}[%
xmin=0,
xmax=15,
xlabel style={font=\normalsize},
xlabel={Timeframe},
ymin=0.3,
ymax=0.9,
ylabel style={font=\normalsize},
ylabel={Mean of Local Gradient Magnitudes},
axis background/.style={fill=white},
title style={font=\normalsize},
title={Average of Local Gradient+Memory Magnitudes},
xmajorgrids,
ymajorgrids,
legend pos=north east,
    legend style={font=\scriptsize},
    ymajorgrids=true,
    grid style=dashed,
]
\addplot [color=blue, dashed]
  table[row sep=crcr]{%
1	0.7599031686\\
2	0.77038070888\\
3	0.6593449804\\
4	0.498023271\\
5	0.4081695358\\
6	0.3630433292\\
7	0.33832931552\\
8	0.3261703886\\
9	0.31863488696\\
10	0.31395564544\\
11	0.31057248912\\
12	0.30805307424\\
13	0.30646215556\\
14	0.30533562872\\
15	0.30448912964\\
};
\addlegendentry{$K=5$, AoG Policy, $\gamma=1$}

\addplot [color=blue]
  table[row sep=crcr]{%
1	0.7585484138\\
2	0.75369341496\\
3	0.6304211022\\
4	0.47753760352\\
5	0.39859661516\\
6	0.35948844572\\
7	0.33945268456\\
8	0.3276091118\\
9	0.32059821688\\
10	0.31610637092\\
11	0.31366434088\\
12	0.3115966364\\
13	0.31031808908\\
14	0.30930791092\\
15	0.30818236236\\
};
\addlegendentry{$K=5$, Genie-Aided, $\gamma=1$}

\addplot [color=red, dashed]
  table[row sep=crcr]{%
1	0.7598759158\\
2	0.79433456788\\
3	0.77739320928\\
4	0.6635645774\\
5	0.52632487664\\
6	0.43784130792\\
7	0.39195609496\\
8	0.36617445832\\
9	0.35065124012\\
10	0.34049392416\\
11	0.33317682972\\
12	0.32739599216\\
13	0.32311032092\\
14	0.31998600728\\
15	0.3171846476\\
};
\addlegendentry{$K=10$, AoG Policy, $\gamma=1$}

\addplot [color=red]
  table[row sep=crcr]{%
1	0.7585481342\\
2	0.77872764588\\
3	0.7488826924\\
4	0.62588947924\\
5	0.50337003008\\
6	0.4293933244\\
7	0.38781544764\\
8	0.36403836796\\
9	0.34876307696\\
10	0.33897823588\\
11	0.33252841384\\
12	0.3282072718\\
13	0.32515980652\\
14	0.32279384756\\
15	0.32066570372\\
};
\addlegendentry{$K=10$, Genie-Aided, $\gamma=1$}

\addplot [color=green, dashed]
  table[row sep=crcr]{%
1	0.75988050088\\
2	0.7971977574\\
3	0.8138315546\\
4	0.78866475808\\
5	0.71047147016\\
6	0.61498382476\\
7	0.54341459504\\
8	0.49150924008\\
9	0.45255519464\\
10	0.42252472408\\
11	0.40109756112\\
12	0.38374708068\\
13	0.36936729652\\
14	0.35882431724\\
15	0.35108809824\\
};
\addlegendentry{$K=20$, AoG Policy, $\gamma=1$}

\addplot [color=green]
  table[row sep=crcr]{%
1	0.75849632652\\
2	0.7894161636\\
3	0.804927751\\
4	0.76580646452\\
5	0.68151761896\\
6	0.59237812184\\
7	0.52317054888\\
8	0.47232919052\\
9	0.43526824692\\
10	0.40998391404\\
11	0.39163791232\\
12	0.37821484352\\
13	0.36797806452\\
14	0.35990269168\\
15	0.35330428448\\
};
\addlegendentry{$K=20$, Genie-Aided, $\gamma=1$}

\end{axis}

\begin{axis}[%
width=5.895in,
height=4.811in,
at={(0in,0in)},
scale only axis,
xmin=0,
xmax=1,
ymin=0,
ymax=1,
axis line style={draw=none},
ticks=none,
axis x line*=bottom,
axis y line*=left,
]
\end{axis}
\end{tikzpicture}%
    \caption{The performance of the AoI policy, as compared with the baselines discussed in Sec. \ref{sec:baselines} for $K=5$. The simulation settings are as in Sec. \ref{sec:DNN Setting}. }
    \label{fig:localgrad_mean}
\end{figure}

\newpage


\begin{thebibliography}{28}
\providecommand{\natexlab}[1]{#1}

\bibitem[{Abramson(1977)}]{Abramson77_PacketBroadcasting}
Abramson, N. 1977.
\newblock The Throughput of Packet Broadcasting Channels.
\newblock \emph{IEEE Trans. on~Commun.}, COM-25(1): 117--128.

\bibitem[{Agarwal et~al.(2018)Agarwal, Suresh, Yu, Kumar, and McMahan}]{FL_DSGD_binomial}
Agarwal, N.; Suresh, A.~T.; Yu, F.~X.; Kumar, S.; and McMahan, B. 2018.
\newblock cp{SGD}: Communication-efficient and differentially-private distributed {SGD}.
\newblock In \emph{32$^{nd}$ Advances in Neural Information Processing Systems (NIPS)}, 7564--7575. Montréal, Canada.

\bibitem[{Alistarh et~al.(2017)Alistarh, Grubic, Li, Tomioka, and Vojnovic}]{alistarh2017qsgd}
Alistarh, D.; Grubic, D.; Li, J.; Tomioka, R.; and Vojnovic, M. 2017.
\newblock {QSGD}: Communication-efficient SGD via gradient quantization and encoding.
\newblock \emph{Advances in Neural Information Processing Systems}, 30.

\bibitem[{{Alistarh} et~al.(2018){Alistarh}, {Hoefler}, {Johansson}, {Konstantinov}, {Khirirat}, and {Renggli}}]{Alistarh2018Spars_FL}
{Alistarh}, D.; {Hoefler}, T.; {Johansson}, M.; {Konstantinov}, N.; {Khirirat}, S.; and {Renggli}, C. 2018.
\newblock The convergence of sparsified gradient methods.
\newblock In \emph{Advances in Neural Information Processing Systems}, 5973--5983.

\bibitem[{Alistarh et~al.(2016, [Online]. Available: http://arxiv.org/abs/1610.02132)Alistarh, Li, Tomioka, and Vojnovic}]{QSGD}
Alistarh, D.; Li, J.; Tomioka, R.; and Vojnovic, M. 2016, [Online]. Available: http://arxiv.org/abs/1610.02132.
\newblock {QSGD}: Randomized quantization for communication-optimal stochastic gradient descent.
\newblock \emph{CoRR}, abs/1610.02132.

\bibitem[{{Bernstein} et~al.(2018){Bernstein}, {Wang}, {Azizzadenesheli}, and {Anandkumar}}]{Bernstein2018signSGC}
{Bernstein}, J.; {Wang}, Y.-X.; {Azizzadenesheli}, K.; and {Anandkumar}, A. 2018.
\newblock sign{SGD}: Compressed optimization for non-convex problems.
\newblock In \emph{Advances in Neural Information Processing Systems}, 560--569.

\bibitem[{Cho, Wang, and Joshi(2020)}]{cho2020client}
Cho, Y.~J.; Wang, J.; and Joshi, G. 2020.
\newblock Client selection in federated learning: Convergence analysis and power-of-choice selection strategies.
\newblock \emph{arXiv preprint arXiv:2010.01243}.

\bibitem[{Dean et~al.(2012)Dean, Corrado, Monga, Chen, Devin, Mao, Ranzato, Senior, Tucker, Yang et~al.}]{dean2012large}
Dean, J.; Corrado, G.; Monga, R.; Chen, K.; Devin, M.; Mao, M.; Ranzato, M.; Senior, A.; Tucker, P.; Yang, K.; et~al. 2012.
\newblock Large scale distributed deep networks.
\newblock \emph{Advances in neural information processing systems}, 25.

\bibitem[{Goetz et~al.(2019)Goetz, Malik, Bui, Moon, Liu, and Kumar}]{goetz2019active}
Goetz, J.; Malik, K.; Bui, D.; Moon, S.; Liu, H.; and Kumar, A. 2019.
\newblock Active federated learning.
\newblock \emph{arXiv preprint arXiv:1909.12641}.

\bibitem[{\hl{McMahan, Brendan and Moore, Eider and Ramage, Daniel and Hampson, Seth and y Arcas, Blaise Aguera}(2017)}]{mcmahan2017communication}
\hl{McMahan, Brendan and Moore, Eider and Ramage, Daniel and Hampson, Seth and y Arcas, Blaise Aguera}. 2017.
\newblock \hl{Communication-efficient learning of deep networks from decentralized data}.
\newblock In \emph{Artificial intelligence and statistics}, 1273--1282. PMLR.

\bibitem[{\hl{Stich, Sebastian U and Cordonnier, Jean-Baptiste and Jaggi, Martin}(2018)}]{stich2018sparsified}
\hl{Stich, Sebastian U and Cordonnier, Jean-Baptiste and Jaggi, Martin}. 2018.
\newblock Sparsified {SGD} with {M}emory.
\newblock \emph{Advances in Neural Information Processing Systems}, 31.

\bibitem[{Karimireddy et~al.(2019)Karimireddy, Rebjock, Stich, and Jaggi}]{karimireddy2019error}
Karimireddy, S.~P.; Rebjock, Q.; Stich, S.; and Jaggi, M. 2019.
\newblock Error feedback fixes sign{SGD} and other gradient compression schemes.
\newblock In \emph{International Conference on Machine Learning}, 3252--3261. PMLR.

\bibitem[{Kaul, Yates, and Gruteser(2012)}]{kaul2012real}
Kaul, S.; Yates, R.; and Gruteser, M. 2012.
\newblock Real-time status: How often should one update?
\newblock In \emph{2012 Proceedings IEEE INFOCOM}, 2731--2735. IEEE.

\bibitem[{Krizhevsky, Nair, and Hinton(2009)}]{cifar10}
Krizhevsky, A.; Nair, V.; and Hinton, G. 2009.
\newblock CIFAR-10 (Canadian Institute for Advanced Research).

\bibitem[{{Li} et~al.(2019){Li}, {Liu}, {Sekar}, and {Smith}}]{Li2019DP_CEFL}
{Li}, T.; {Liu}, Z.; {Sekar}, V.; and {Smith}, V. 2019.
\newblock Privacy for free: Communication efficient learning with differential privacy using sketches.
\newblock \emph{Available: https://arxiv.org/abs/1911.00972}.

\bibitem[{Li et~al.(2020)Li, Sahu, Zaheer, Sanjabi, Talwalkar, and Smith}]{li2020federated}
Li, T.; Sahu, A.~K.; Zaheer, M.; Sanjabi, M.; Talwalkar, A.; and Smith, V. 2020.
\newblock Federated optimization in heterogeneous networks.
\newblock \emph{Proceedings of Machine learning and systems}, 2: 429--450.

\bibitem[{Mahmoudi, Ghadikolaei, and Fischione(2020)}]{afsane2020}
Mahmoudi, A.; Ghadikolaei, H.~S.; and Fischione, C. 2020.
\newblock Cost-efficient Distributed optimization In Machine Learning Over Wireless Networks.
\newblock In \emph{ICC 2020 - 2020 IEEE International Conference on Communications (ICC)}, 1--7.

\bibitem[{Ribero and Vikalo(2020)}]{ribero2020communication}
Ribero, M.; and Vikalo, H. 2020.
\newblock Communication-efficient federated learning via optimal client sampling.
\newblock \emph{arXiv preprint arXiv:2007.15197}.

\bibitem[{{Seide} et~al.(2014){Seide}, {Fu}, {Droppo}, {Li}, and {Yu}}]{seide2014onebitSGD}
{Seide}, F.; {Fu}, H.; {Droppo}, J.; {Li}, G.; and {Yu}, D. 2014.
\newblock 1-bit stochastic gradient descent and its application to data-parallel distributed training of speech {DNN}s.
\newblock In \emph{INTERSPEECH}, 9850--9861.

\bibitem[{Shai~{Shalev-Shwartz} and {Zhang}(2010)}]{Shalev-Shwartz2010FL_CE}
Shai~{Shalev-Shwartz}, N.~S.; and {Zhang}, T. 2010.
\newblock Trading accuracy for sparsity in optimization problems with sparsity constraints.
\newblock \emph{SIAM J. Optimization}.

\bibitem[{Simonyan(2014)}]{simonyan2014very}
Simonyan, K. 2014.
\newblock Very deep convolutional networks for large-scale image recognition.
\newblock \emph{arXiv preprint arXiv:1409.1556}.

\bibitem[{Stich and Karimireddy(2020)}]{stich2020error}
Stich, S.~U.; and Karimireddy, S.~P. 2020.
\newblock The error-feedback framework: Better rates for sgd with delayed gradients and compressed updates.
\newblock \emph{Journal of Machine Learning Research}, 21: 1--36.

\bibitem[{{Wang} et~al.(2018){Wang}, {Sievert}, {Liu}, {Charles}, {Papailiopoulos}, and {Wright}}]{Wang2018Spars_FL}
{Wang}, H.; {Sievert}, S.; {Liu}, S.; {Charles}, Z.; {Papailiopoulos}, D.; and {Wright}, S. 2018.
\newblock Atomo: Communication-efficient learning via atomic sparsification.
\newblock In \emph{Advances in Neural Information Processing Systems}, 9850--9861.

\bibitem[{Wang et~al.(2024)Wang, Ma, Mashhadi, Foh, Tafazolli, and Ding}]{wang2024convergenc}
Wang, K.; Ma, Y.; Mashhadi, M.~B.; Foh, C.~H.; Tafazolli, R.; and Ding, Z. 2024.
\newblock Convergence Acceleration in Wireless Federated Learning: A Stackelberg Game Approach.
\newblock arXiv:2209.06623.

\bibitem[{Wangni et~al.(2018)Wangni, Wang, Liu, and Zhang}]{wangni2018gradient}
Wangni, J.; Wang, J.; Liu, J.; and Zhang, T. 2018.
\newblock Gradient sparsification for communication-efficient distributed optimization.
\newblock \emph{Advances in Neural Information Processing Systems}, 31.

\bibitem[{Wen et~al.(2017)Wen, Xu, Yan, Wu, Wang, Chen, and Li}]{wen2017terngrad}
Wen, W.; Xu, C.; Yan, F.; Wu, C.; Wang, Y.; Chen, Y.; and Li, H. 2017.
\newblock Terngrad: Ternary gradients to reduce communication in distributed deep learning.
\newblock \emph{Advances in neural information processing systems}, 30.

\bibitem[{Yang et~al.(2019)Yang, Liu, Quek, and Poor}]{yang2019scheduling}
Yang, H.~H.; Liu, Z.; Quek, T.~Q.; and Poor, H.~V. 2019.
\newblock Scheduling policies for federated learning in wireless networks.
\newblock \emph{IEEE transactions on communications}, 68(1): 317--333.

\bibitem[{Yates et~al.(2021)Yates, Sun, Brown, Kaul, Modiano, and Ulukus}]{yates2021age}
Yates, R.~D.; Sun, Y.; Brown, D.~R.; Kaul, S.~K.; Modiano, E.; and Ulukus, S. 2021.
\newblock Age of information: An introduction and survey.
\newblock \emph{IEEE Journal on Selected Areas in Communications}, 39(5): 1183--1210.

\end{thebibliography}
\end{document}